# Design and Implementation a 8 bits Pipeline Analog to Digital Converter in the Technology 0.6 μm CMOS Process.

**Eri Prasetyo\*, Dominique Ginhac\*\* , Michel Paindavoine\*\***

*\* Gunadarma University, Indonesia*
*\*\*Laboratoire LE2I - UMR CNRS 5158*
*Université de Bourgogne*
*21078 Dijon Cedex – FRANCE*

*E-mail: (dginhac,paindav)@u-bourgogne.fr,eri@staff.gunadarma.ac.id*

**ABSTRACT**

This paper describes a 8 bits, 20 Msamples/s pipeline analog-to-digital converter implemented in 0.6 μm CMOS technology with a total power dissipation of 75.47 mW. Circuit techniques used include a precise comparator, operational amplifier and clock management. A switched capacitor is used to sample and multiplying at each stage. Simulation a worst case DNL and INL of 0.75 LSB. The design operate at 5 V dc. The ADC achieves a SNDR of 44.86 dB.
*keywords : pipeline, switched capacitor, clock management*

## 1. Introduction

This ADC is designed to convert of pixels analogic value to digital value in the my image CMOS sensor design. The goal from ADC design is to run in the video rate and it must have small size in the layout design.

Most traditional designs of video-rate analog to digital converters (ADC's) of 8 bit resolution are implemented through FLASH architectures and bipolar technologies[1]. For this application, it require power dissipation 250 mW; its conversion rate was was limited to 15 Msamples/s[1]. In recent years, pipelined switched - capacitor topologies have emerged as approach to implementing power-efficient nyquist-rate ADC that have medium-to-high resolution at medium-to-high conversion rates[2],[3]. A switched capacitor circuit is implemented with the Sample an Hold (SH) amplifies the signal for finish conversion down the pipeline.

The paper presents a 8 bit pipeline ADC which operating at a 5 V dc that achieve sample rate environ 20 Msamples/s and power dissipation 75.47 mW. An experimental prototype of converter has been implemented in 0.6 æm.

## 2. One-Bit Per Stage Pipeline Architecture

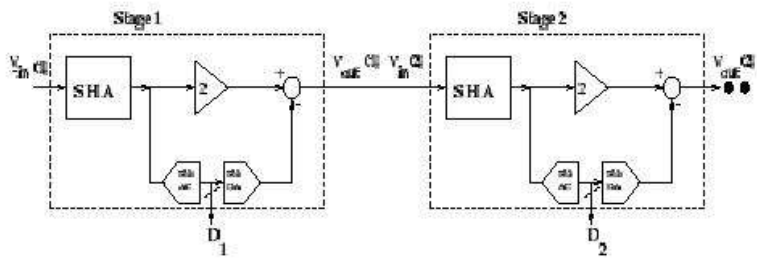

Figure 1. One bit/ stage architecture

Makalah ada di prosiding ISSM05 , Paris, 30th September – 1st October 2005

Figure 1 shows block diagram of an ideal N-stage, 1-bit per stage pipelined A/D converter. Each stage contributes a single bit to digital output. The most significant bits are resolved by first stage in the pipeline. The result of stage is passed on the next stage where the cycle is repeated. A pipeline stage is implemented by the conventional switched capacitor ( SC ), it is shown at figure 2[4].

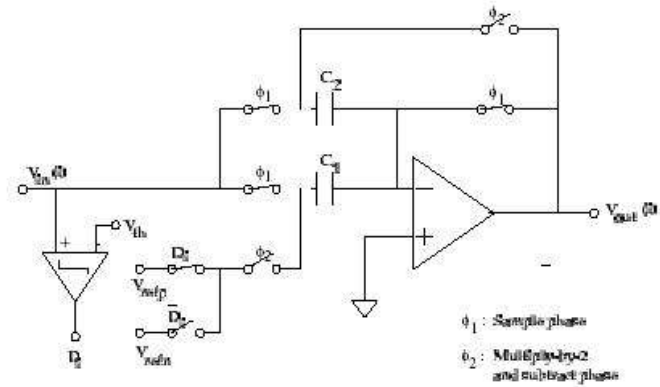

Fiure. 2. Scheme of switched capacitor pipelined A/D converter

V*refp* is the positive reference voltage and V*refn* is a negative reference voltage. Each stage consists capacitor C1, C2 , an operational amplifier and a comparator. Value of *C1* and *C2* are equal in my design. Each stage operates in two phase, a sampling phase and a multiplying phase.

During the sampling phase $\phi 1$ , the comparator produces a digital output *Di*. *Di* is 1 if *Vin* > *Vth* and *Di* is 0 if *Vin* < *Vth,* where *Vth* is the threshold voltage defined midway between *Vrefp* and *Vrefn.* During multiplying phase, C2 is connected to the output of the operational amplifier and C1 is connected to either the reference voltage *Vrefp* or *Vrefn*, depending on the bit value *Di*. If *Di* = 1, C1 is connected to *Vrefp*, resulting in the resedu ( Vout ) is :

$$Vout(i) = 2 \times Vin(i) - Di.Vrefp \qquad (1)$$

Otherwise, C1 is connected to *Vrefn*, giving an output voltage :

$$Vout(i) = 2 \times Vin(i) - \check{D}.Vrefn \qquad (2)$$

**3. Comparator**

Precision comparator is implemented to each stage of the ADC. We prefer to use precision comparator then digital correction to minimize offset error of comparator and better output of ADC.

This comparator consists of three blocks:preamplifier, decision circuit and output buffer. first block is the input preamplifier which the circuit is a differential amplifier with active loads. The size of transistors m_2 and m_3 are set by considering the diff-amp transconductance and the input capacitance. Second block is a positive feedback or decision circuit, it is the heart of the comparator. The circuit uses positive feedback from the cross gate connection of m_11 and m_12 to increase the gain of the decision element. Third stage is output buffer, it is to convert the output of the decision circuit into a logic signal. The inverter (m_20 and m_21) is added to isolate any load capacitance from the self biasing differential amplifier. The complete circuit of comparator is shown in figure 3.

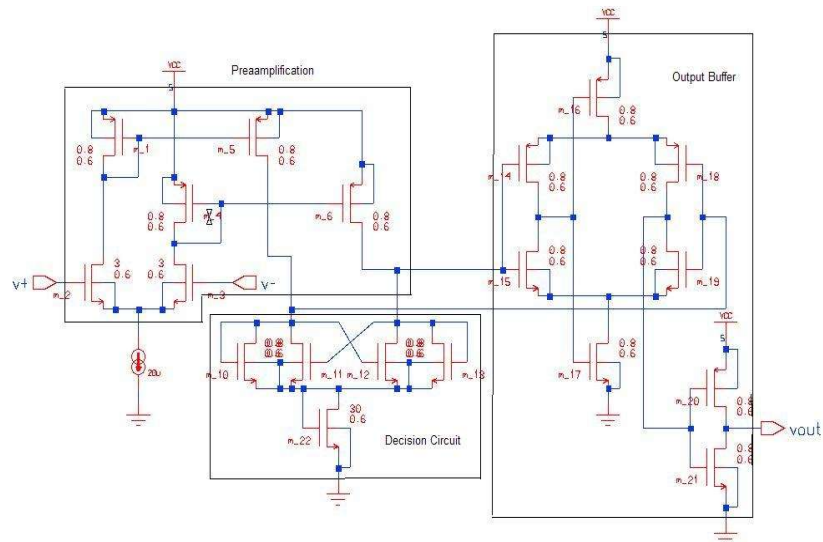

Figure 3. The comparator circuit

## 4. Operational Amplifier

In this pipeline ADCs, operational amplifier is very important to get accurately result. We used an operational transconductance amplifier which has a gain of approximately 55 dB for a bias current of 2.5 µA with Vdd = 5 V and Vss = -5 V. A value of loading capacitor is 0.1 Pf. The complete circuit is shown in figure 4. Transistors m_1_1_1 and m_1_1 functions as a constant current source, and transistors m_1, m_2 and m_3 functions as two current mirror 'pairs'. The transistors m_4, m_5, m_6 and m_7 are the differential amplifier. Transistor m_9 is an output amplifier stage. In the simulation, we got the resultat for phase margin (PM) was -145 degre, A gain was 55 dB and Gain bandwidth product was 800 MHz. A power dissipation mesured of 10.825 mW.

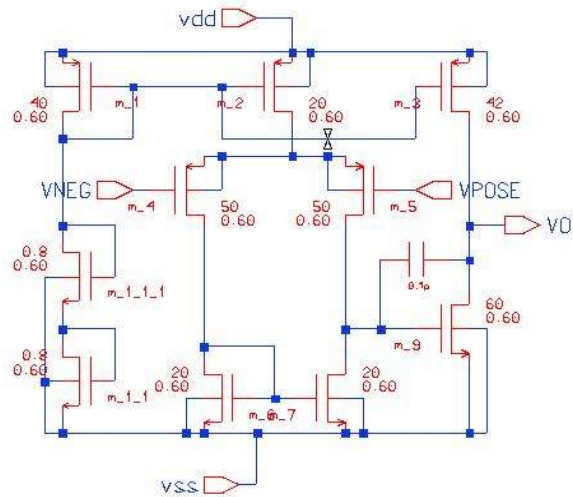

Figure 4. Transconductance OP-AMP

## 5. Clock Management

In the design pipeline A/D converter use latch technique is used to hold active condition at multiplying ø2 ( phi2 ) and non active condition at sampling ø1 ( phi1 ) until next stage begin to execute sampling phase. This purpose to keep the output voltage of residu from before stage conformity at input next stage.

The clock management system use counter to count some clock to active the address decoder from each stage. Signal output decoder active reset signal so the clock management begin working. This work is begun from early address to last address. Ending of address decoder, a stop decoder give reset signal stopping activity pipeline ADC's. The complete circuit is shown at figure 5.

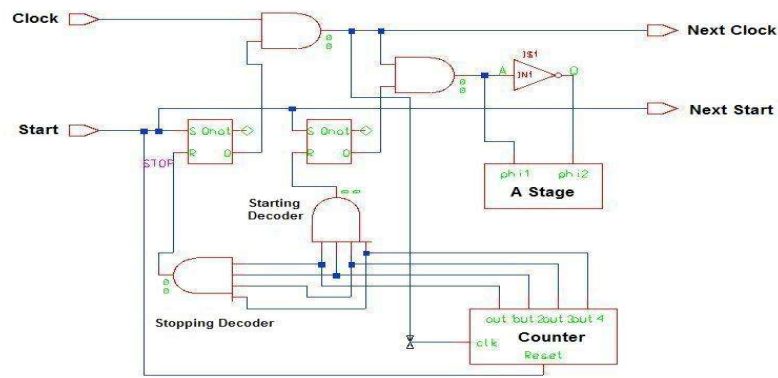

Figure 5. The circuit of clock management

## 6. Result

One stage A/D converter layout was estimated to occupy about 174 μm x 89 μm, it is seen at figure 6.

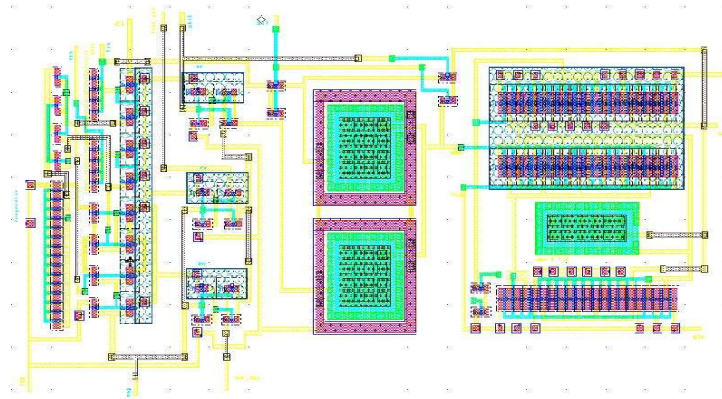

Figure 6. One stage A/D converter layout

Figure 7 shows the dc linearity of the ADC at conversion rate of 20 Msamples/s. In the figure 7(a), the CODE is plotted versus integral nonlinearity (INL) value and figure 7(b), the CODE is plotted versus differntial nonlinearity (DNL). Note that since each simulation lasted 20 minutes, only 25 code were tested. As shown, the worst INL is less than 0.8 LSB; the DNL is less then 0.8 LSB.

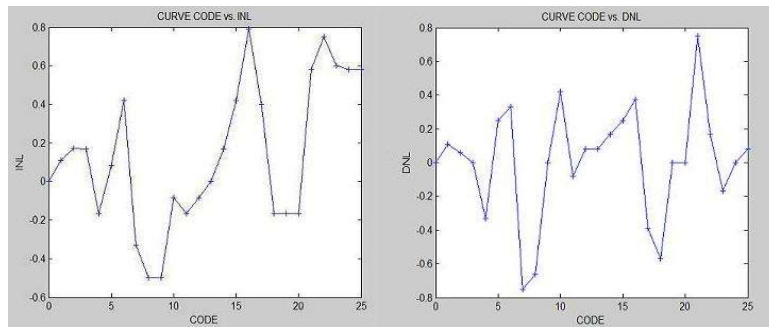

Figure 7. (a) Curve Code vs INL and (b) Curve Code vs DNL

Figure 8 shows the output of Fast Fourier transform (FFT) on a blocks of 1024 consecutive codes. The conversion rate is 20 Msamples/s, and the input is a full scale sine wive at 10 Mhz. From curve FFT, The signal-to-noise plus distorsion ratio (SNDR) is obtained about 44.86 dB. The effective number of bits (ENOB) is calculated environ 7.2 bits.

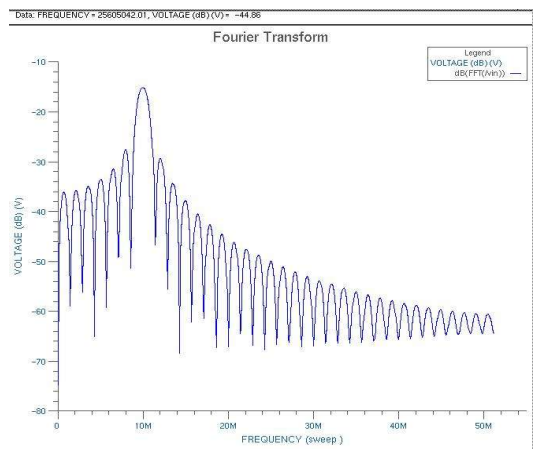

Figure 8. Curve FFT

## 7. Conclusion

The pipeline ADC 8 bits, 20 Msamples/s was implemented in 0.6 μm technology with total power dissipation 75.47 mW. Refer to result of experiment, the ADC can be implemented for video rate application.

The system use clock management to manage data conversion so that the system is simple and have good precision.

**References**
[1] S. H. Lewis and H. Scott Fetternan and George F. Gross jr. and R. Ramachandran and T. R. Vismanathan, "10-b 20 Msamples/s analog-to-digital converter," *Journal of IEEE Solid State Circuit,* vol. 27, pp. 351-358, March 1992.
[2] Dwight U. Thomson and Bruce A. Wooley, "A 15-b pipelined CMOS floating point A/D converter, " *Journal of IEEE Solid State Circuit,* vol. 36, no. 2, February 2001.
[3] Timothy M. Hancock and Scott M. Pernia and Adam C. Zeeb, " A Digitally corrected 1.5 bits/stage low-power 80 Ms/s 10-bits pipelined ADC, " EECS 589-02 University of Minchigan Tech. rep., December 2002.
[4] R. Samer and Jan Van der Speigel and K. Nagaraj, "Background digital error correction technique for pipeline ADC, " *IEEE,* 2001.
[5] A. Shabra and Hae-Seung Lee, " Oversampled pipelined A/D converter with mismatch shaping," *Journal of IEEE Solid State Circuit,* vol. 37, no. 5, May 2002.